%% file: root.tex
\documentclass[a4paper,conference]{IEEEtran}

\input{header}
\begin{document}
\title{Rethinking Experience Replay:\\a Bag of Tricks for Continual Learning}
\author{
    \IEEEauthorblockN{Pietro Buzzega\IEEEauthorrefmark{1}, Matteo Boschini\IEEEauthorrefmark{1}, Angelo Porrello and Simone Calderara}
    \IEEEauthorblockA{University of Modena and Reggio Emilia, Italy}
    Email: {\{pietro.buzzega, matteo.boschini, angelo.porrello, simone.calderara\}@unimore.it}
}
\maketitle
% As a general rule, do not put math, special symbols or citations
% in the abstract
\begin{abstract}
\input{abs}
\end{abstract}
\IEEEpeerreviewmaketitle
\section{Introduction} \label{sec:intro}
{\let\thefootnote\relax\footnote{{\IEEEauthorrefmark{1} indicates \textit{equal contribution}}}}
Deep Neural Networks represent a valid tool for classification tasks, showing excellent performance on a variety of domains. However, this holds when all training data are immediately available and identically distributed, a condition that is hard to find outside an artificial environment. In a practical application, new classes may emerge later in the stream: simply fine-tuning on them would disrupt the previously acquired knowledge quickly. This is known as Catastrophic Forgetting problem~\cite{mccloskey1989catastrophic}, arising whenever the data stream faces a shift in its distribution. Continual Learning (CL) algorithms aim at learning from that stream, retaining the old knowledge while relying on bounded computational resources and memory footprint~\cite{rebuffi2017icarl}.

Researchers and practitioners model the aforementioned shift through various evaluation protocols, which typically involve a sequence of different classification problems (tasks). Given the categorization presented in~\cite{hsu2018re,van2019three}, we primarily focus on Class Incremental Learning (Class-IL) due to its challenging nature~\cite{farquhar2018towards}. In such a setting, a dataset (\textit{e.g.}\ MNIST) is divided into class-based partitions (\textit{e.g.} 0 vs 1, 2 vs 3 etc.), each of which is considered a separate task. The model -- which observes tasks in a sequential fashion -- ought to learn new classes without impairing its knowledge about the old ones. Since no information about the task is given at inference time, the test phase requires the network to infer the right class label among all those seen. Therefore, this precludes the adoption of a multi-head classification layer.

\begin{figure}[t]
\centering
\includegraphics[width=0.48\textwidth]{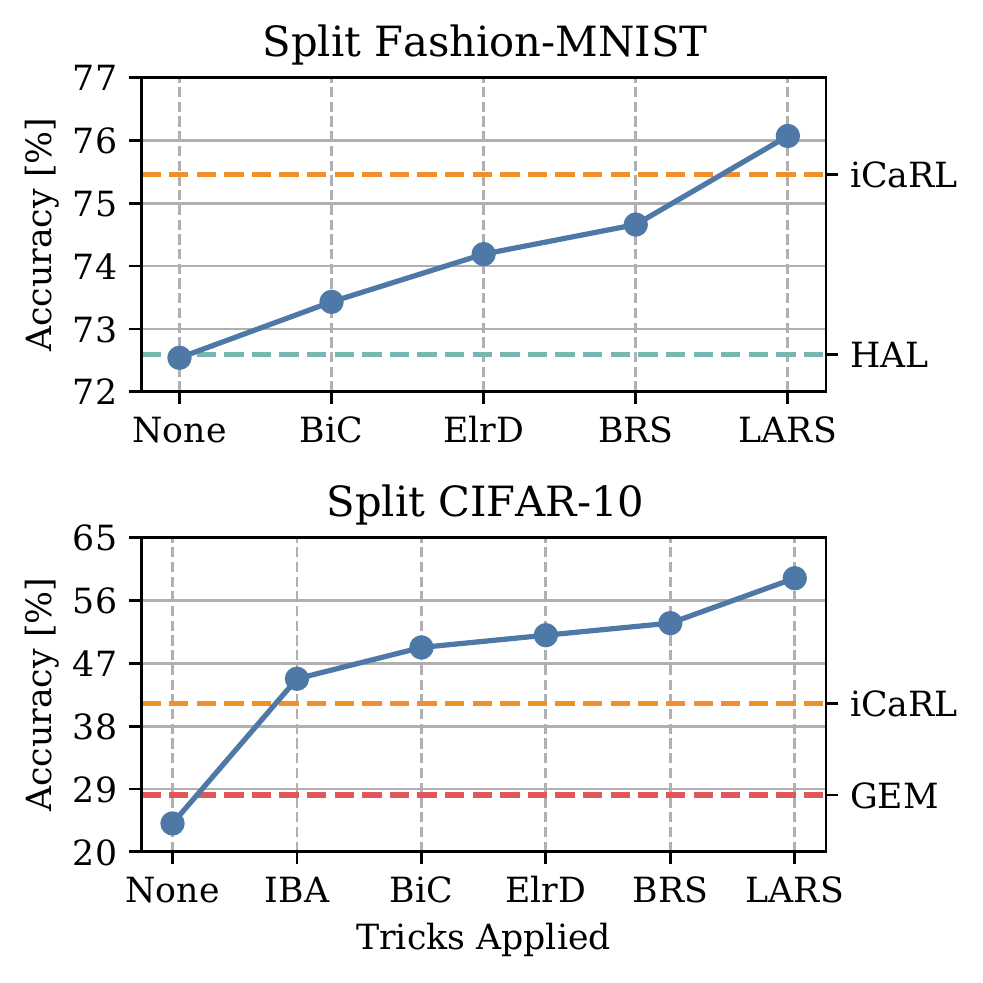}
\vspace{-1em}
\caption{Accuracy on different datasets as more training tricks are applied to Experience Replay, with a replay buffer containing $200$ examples. Accuracy of competitor methods are also reported \textit{(best in color)}.}
\label{fig:perf_tricks}
\end{figure}

Current approaches can be broadly categorized in two branches: on the one hand, prior-based methods \cite{kirkpatrick2017overcoming, zenke2017continual, chaudhry2018riemannian, rusu2016progressive,mallya2018packnet, serra2018overcoming} estimate the importance of each weight for past tasks and protect these from later updates; on the other hand, rehearsal-based approaches \cite{ratcliff1990connectionist, lopez2017gradient, riemer2018learning, rebuffi2017icarl} rely on a subset (buffer) of old data to avoid interference.
Both these families of solutions require additional computations and a large memory footprint, but rehearsal-based methods have shown to be far better for incrementally learning new classes~\cite{farquhar2018towards}.
Among them, many recent works rely on complex techniques applied on top of rehearsal~\cite{riemer2018learning, aljundi2019gradient, aljundi2019online}. Nevertheless, computing a regularization term over a buffer of past examples usually entails a non-negligible overhead, as the network needs to make additional feed-forward passes.

Here, we follow the principle of Occam's razor and step back to Experience Replay~\cite{ratcliff1990connectionist} (ER), which simply consists of interleaving past examples with the current training batch. Thanks to its straightforward formulation, it is robust at the cost of a slight increase in computational requirements. In this paper, we work on alleviating some problems encountered by naive replay in the Class-IL setting. In more detail, we address the following issues:

\begin{enumerate}[label=(\alph*)]
    \item Rehearsal methods could experience overfitting, as they repeatedly optimize the examples stored in a relatively small buffer\footnote{An early-sampled item in the Split CIFAR-10 protocol (buffer size $500$; replay batch size $32$) is replayed approximately $5000$ times.};
    \item Incrementally learning a sequence of classes implicitly biases the network towards newer ones, making the performance unbalanced in favor of the latest task encountered~\cite{wu2019large, hou2019learning};
    \item The memory buffer is commonly populated by sampling random items from the training stream, with the aim of obtaining an i.i.d.\ distribution~\cite{riemer2018learning,chaudhry2019on}. While this is generally valid, there are failure cases. As an example, it may leave out some classes when the buffer is small.
\end{enumerate}

We illustrate how few modifications (a bag of \textit{tricks}) not only mitigate the above-mentioned issues, but also make the simplest rehearsal baseline outperform the SOTA. Our main contributions are:
\begin{itemize}
    \item a collection of five modifications that can be easily applied to CL methods to improve their performance;
    \item experiments showing that ER -- when equipped with these \textit{tricks} -- outperforms state-of-the-art rehearsal methods on four distinct Class-IL experimental settings, thus providing a new reliable baseline for any practitioner approaching Continual Learning;
    \item further analysis clarifying how the \textit{tricks} affect the accuracy of ER and their applicability to other methods.
\end{itemize}
In light of the remarkable increase in performance that results from the application of our proposed \textit{tricks}, we hope that this work will constitute a valuable quick reference as well as a stepping stone for the design of more accurate CL approaches.
 
\section{Related Works} \label{sec:related}

After the first formulation of catastrophic forgetting in ANNs~\cite{mccloskey1989catastrophic}, early works dating back to the 1990s~\cite{ratcliff1990connectionist, robins1995catastrophic} proposed a straightforward solution to it, that is storing previously seen examples in a memory buffer and later interleaving them with new training batches. We refer the reader to Sec.~\ref{sec:baseline} for a detailed description of this simple, yet still very effective method, generally known as Experience Replay (ER).

In the 2010s, the advent of Deep Neural Networks sparked renewed interest in Continual Learning~\cite{goodfellow2013empirical} leading to a very prolific production of works trying to counter forgetting. These contributions can be split into two categories: those that store parameters and those that store past examples.

The first group encompasses all methods that save past network configurations to prevent drifting away from them. In~\cite{rusu2016progressive}, \textit{Rusu et al.}\ delegate distinct tasks to distinct neural networks, limiting interference by design. Although this is a very effective strategy, it must be noted that the cost for storing entire new models is very high (it grows linearly w.r.t.\ the number of tasks).~\cite{mallya2018packnet, serra2018overcoming, fernando2017pathnet} also suggest similar approaches that adapt the model's architecture to multi-task settings. Alternatively,~\cite{li2017learning, kirkpatrick2017overcoming, zenke2017continual, chaudhry2018riemannian, wu2019large, hou2019learning} keep a model checkpoint and use it to formulate an additional loss term to prevent excessive weight alterations in later tasks.

On the opposite side of the spectrum, methods storing past examples relate to ER, in that they also tackle forgetting by collecting exemplars seen in previous training phases. \textit{Rebuffi et al.}\ propose iCaRL~\cite{rebuffi2017icarl}, an approach that classifies new exemplars by finding the nearest mean representation of past elements in an incrementally learned feature space. GEM~\cite{lopez2017gradient} and its more efficient version A-GEM~\cite{chaudhry2018efficient} showcase another non-obvious use of memories, as they define inequality constraints to avoid the increase in the losses w.r.t.\ previous tasks while allowing for their decrease.

Although -- quite remarkably -- none of the mentioned papers draw any experimental comparison with it, Experience Replay still appears to be a strong solution to catastrophic forgetting. This is corroborated by a line of very recent works~\cite{riemer2018learning, chaudhry2019on, aljundi2019gradient, aljundi2019online, riemer2019scalable, chaudhry2020using, buzzega2020dark} that argue for its superior effectiveness in comparison to other methods and propose various extensions to it. Meta-Experience Replay~\cite{riemer2018learning} notably complements replay with meta-learning techniques to maximize transfer and minimize interference. The authors of~\cite{aljundi2019gradient} propose an optimized strategy for choosing what to store in the memory buffer, while~\cite{aljundi2019online} explores an alternative policy for sampling from it. On the other hand,~\cite{riemer2019scalable} suggests adopting a Variational Autoencoder as a compression mechanism to maximize the efficiency of storing replay exemplars. Most recently, HAL~\cite{chaudhry2020using} equips ER with a regularization term anchoring the network's responses towards data-points learned from classes of previous tasks and DER~\cite{buzzega2020dark} shows that simply replaying network responses instead of ground truth labels produces an even stronger baseline than ER.

It is also worth mentioning that other works~\cite{shin2017continual, lavda2018continual} exploit generative models in place of the memory buffer. On the one hand, this favorably allows sampling data-points from the distributions underlying the old tasks. On the other hand, it requires training the generator online, thus giving rise to a chicken and egg problem. Furthermore, as~\cite{shin2017continual} regards ER as an upper bound for the proposed method, it appears that the former is currently a more effective and viable solution.

In the light of these considerations, we set out to prove that ER can perform significantly better than other approaches when equipped with the tricks described in Sec.~\ref{sec:tricks}.

\section{Baseline Method}
\label{sec:baseline}

A Class Incremental classification problem requires training a learner $f_{\theta}$ to classify samples $x$ that belong to one class $y$ out of a given dataset $\mathcal{D}$. This corresponds to minimizing the following loss term:
\begin{equation}
    \mathcal{L} = {\mathds{E}_{(x, y) \sim \mathcal{D}}\big[\ell(y, f_{\theta}(x))\big]}.
    \label{eq:obj}
\end{equation}

In a CL scenario, the dataset is divided into $N_t$ tasks, during each of which $t\in\{1,...,N_t\}$ only examples belonging to a specific partition of classes $\mathcal{D}_t$ are shown.
As the classifier only observes one portion of the dataset at a time, it cannot optimize Eq.~\ref{eq:obj} directly. Therefore, it must devise a strategy to approximate it. \textbf{Experience Replay} addresses this issue by storing exemplars and labels from previous tasks in a replay buffer $\mathcal{B}$. During each training step, it merges some of these items with the current batch: consequently, the network rehearses past tasks as it learns current data. This amounts to optimizing the following loss term as a surrogate of Eq.~\ref{eq:obj}:
\begin{equation}
    \mathcal{L}' = \mathds{E}_{(x, y) \sim \mathcal{D}_t}\big[\ell(y, f_{\theta}(x))\big] + \mathds{E}_{(x, y) \sim \mathcal{B}}\big[\ell(y, f_{\theta}(x))\big].
    \label{eq:obj_er}
\end{equation}

This practical solution only introduces two additional hyper-parameters to $f_\theta$, namely the replay buffer size $|\mathcal{B}|$ and the number of elements that we draw from it at each step. In this work, we initially assume to update the buffer with the \textit{reservoir} sampling strategy~\cite{vitter1985random} (see Alg.~\ref{alg:balancoir}), as proposed in~\cite{riemer2018learning}. This guarantees that each input exemplar has the same probability $|\mathcal{B}|/|\mathcal{D}|$ of entering the replay buffer. We prefer this solution both to \textit{herding}~\cite{rebuffi2017icarl} and the \textit{class-wise FIFO}~\cite{chaudhry2019on} strategies (a.k.a.\ \textit{ring} buffer). Unlike \textit{reservoir}, the former needs to retain the entire training set of each task. Conversely, the latter does not exploit the whole memory and is more likely to overfit (as pointed out in~\cite{chaudhry2019on}).

\section{Training Tricks}
\label{sec:tricks}
In this section, we discuss some issues that ER encounters in the Class-IL setting and propose effective tricks to mitigate them. On the one hand, Sec.~\ref{subsec:mem_aug}, \ref{subsec:balancoir} and~\ref{subsec:lossoir} describe improvements to the replay buffer and can, therefore, easily extend to other rehearsal-based methods. On the other hand, the tricks described in Sec.~\ref{subsec:bic} and~\ref{subsec:lr_d} are more general and applicable to any Class-IL method.

\begin{figure}[t]
\centering
\includegraphics[width=0.48\textwidth]{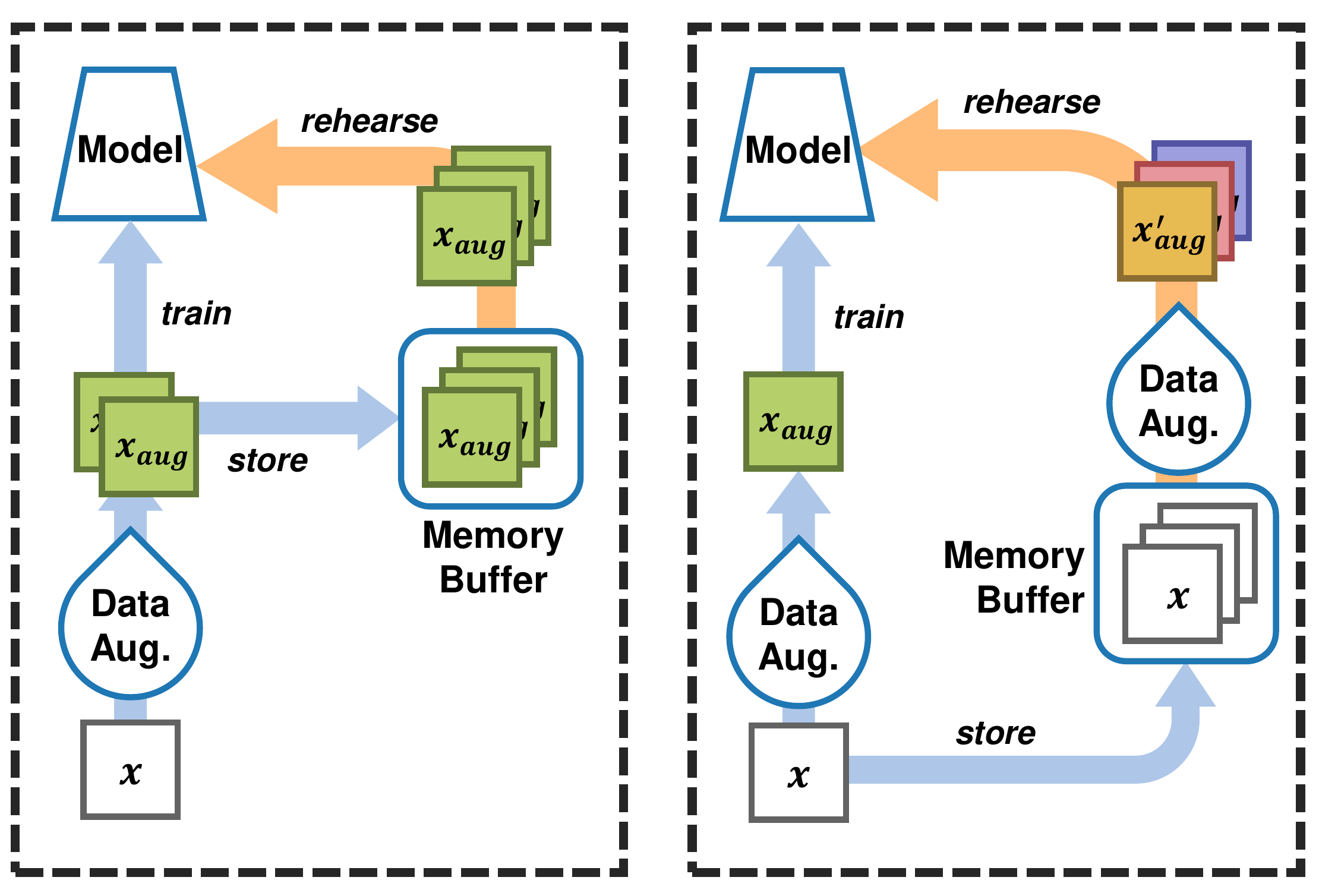}
\\
\begin{tabularx}{.47\textwidth}{XX}
\centering (a)&\centering (b)\\
\end{tabularx}
\vspace{-1.5em}
\caption{Graphical comparison between rehearsal on augmented examples (a) and Independent Buffer Augmentation (b) \textit{(best in color)}.}
\label{fig:iba}
\end{figure}

\subsection{Independent Buffer Augmentation (IBA)}
\label{subsec:mem_aug}

Data augmentation is an effective strategy for improving the generalization capabilities of a Deep Network~\cite{wong2016understanding}. When dealing with Continual Learning scenarios, one can apply data augmentation on the input stream of data (\textit{i.e.} the examples shown to the net during each task). However, for a rehearsal method, replayed exemplars constitute a significant portion of the overall training input. This could pose a serious risk of overfitting the memory buffer, which we address through Independent Buffer Augmentation (IBA): in addition to the regular augmentation performed on the input stream, we store examples not augmented in the memory buffer; this way, we can augment them independently when drawn for later replay. By so doing, we minimize overfitting on the memory and introduce additional variety in the rehearsal examples.

While adopting this simple expedient could seem a no-brainer, its application in literature should not be taken for granted. As an example, the CL methods implemented in the codebase of~\cite{lopez2017gradient}\footnote{\href{https://github.com/facebookresearch/GradientEpisodicMemory}{https://github.com/facebookresearch/GradientEpisodicMemory}} and~\cite{aljundi2019gradient}\footnote{\href{https://github.com/rahafaljundi/Gradient-based-Sample-Selection}{https://github.com/rahafaljundi/Gradient-based-Sample-Selection}} store the augmented examples in the memory buffer and re-use them as-they-are, as illustrated in Fig.~\ref{fig:iba}(a). On the contrary, we remark that it is much more beneficial to show the model replay examples that undergo distinct transformations, as shown in Fig.~\ref{fig:iba}(b). 

\subsection{Bias Control (BiC)}
\label{subsec:bic}
Given the sequential nature of the Class-IL setting, the network's predictions end up showing bias towards the current task. Indeed, a single-head classifier is less prone to predicting classes found in prior tasks than those learned just before testing. Such bias is linked to the whole model and not exclusively to its final classification layer: consequently, the trivial solution of zeroing the latter is not beneficial.

This imbalance problem is analyzed by both \textit{Hou et al.}\ in~\cite{hou2019learning} and \textit{Wu et al.}\ in~\cite{wu2019large}. The former addresses this issue structurally by devising a specific margin-ranking loss term aimed keeping representations from different tasks separated. However, the latter work proposes a much simpler and modular solution, which we also apply here: the addition of a simple Bias Correction Layer to the model. This layer consists in a linear model with two parameters $\alpha$ and $\beta$ that compensates the $k^\text{th}$ output logit $o_k$ as follows.
\begin{equation}
    q_k = \begin{cases}
      \alpha \cdot o_k + \beta & \text{if $k$ was trained in the last task} \\
      o_k & \text{otherwise}
    \end{cases}
\end{equation}
Such a layer is applied downstream of the classifier to yield the final output at test time. Thanks to its small size, it can be easily trained at the end of each task by leveraging a limited amount of exemplars. Importantly, while \cite{wu2019large} employs a separate validation set for this purpose, we simply exploit the same replay buffer we use for rehearsal methods. Parameters $\alpha$ and $\beta$ are optimized through the cross-entropy loss, as follows:
\begin{equation}
    \mathcal{\ell}_{\text{BiC}} = - \sum_{k} \delta_{y=k} \operatornamewithlimits{log}[\operatornamewithlimits{softmax}(q_k)].
\end{equation}

We agree with the authors of~\cite{wu2019large} on the effectiveness of this simple linear model to counter the above-mentioned bias.
\subsection{Exponential LR Decay (ELRD)}
\label{subsec:lr_d}
Arguably, the best way to preserve previous knowledge is not to learn anything new. To this aim, we propose to decrease the learning rate progressively at each iteration; we found exponential decay particularly effective. Exponential-based rules for decaying the learning rate were early introduced in literature to speed up the learning process~\cite{an2017exponential,li2019exponential}. CL algorithms that exploit this technique~\cite{kirkpatrick2017overcoming,rebuffi2017icarl} do so in a task-wise manner (namely, the schedule starts again at the beginning of each task). Differently, we point out that decreasing the learning rate for the whole duration of the training relieves catastrophic forgetting. We thus recommend to compute the learning rate for the $j^\text{th}$ example as follows:
\begin{equation}
    lr_j = lr_0 \cdot {\gamma}^{N_{ex}},
\end{equation}
where $N_{ex}$ is the number of input examples seen so far, $lr_0$ indicates the initial learning rate for training and $\gamma$ is a hyper-parameter tuned to make the learning rate approximately $\nicefrac{1}{6}$ of the initial value at the end of the training.

It is worth noting that decreasing the learning rate yields an additional regularization objective, which penalizes weights change between subsequent steps. In Continual Learning, a family of non-rehearsal approaches~\cite{kirkpatrick2017overcoming, zenke2017continual, chaudhry2018riemannian} rely on this concept, applying a loss term to prevent the same kind of interference. However, ELrD does not produce the additional overhead that characterizes these methods.

\begin{algorithm}[t]
    \caption{Balanced Reservoir Sampling}
    \label{alg:balancoir}
    \begin{algorithmic}[1]
      \STATE {\bfseries Input:} exemplar $(x,y)$, replay buffer $\mathcal{B}$, 
      \STATE \hspace{2.70em} number of seen examples $N$.
      \IF{$|\mathcal{B}| > N$}
            \STATE $\mathcal{B}[N]\gets (x,y)$
      \ELSE
            \STATE $j \gets \operatornamewithlimits{RandInt}([0, N])$
            \IF {$j < |\mathcal{B}|$}
                \vspace{5.5mm}
                \STATE \tikzmark{start6}$\mathcal{B}[j]\gets (x,y)$  \tikzmark{end6} \\ \vspace{-9.4mm} \hspace{-0.7em} \textbf{Reservoir Sampling} 
                \vspace{3.2em}
                \vspace{1mm}
                \STATE $\tilde{y} \gets \small{\argmax} \ \operatornamewithlimits{ClassCounts}(\mathcal{B},y)$
                \STATE\tikzmark{start7}$k\gets  \operatornamewithlimits{RandChoice}(\{\tilde{k}; \mathcal{B}[\tilde{k}] = (x,y), y = \tilde{y}\})$ \tikzmark{end7} 
                \STATE $\mathcal{B}[k]\gets (x,y)$ \\ \vspace{-5em} \hspace{-0.7em} \textbf{Balanced Reservoir Sampling} \vspace{4em}
                \vspace{1.2mm}
            \ENDIF
      \ENDIF
    \vspace{-1.2em}
    \end{algorithmic}
\TextboxReservoir[0]{start6}{end6}{}
\TextboxBalancedReservoir[0]{start7}{end7}{}
\end{algorithm}

\subsection{Balanced Reservoir Sampling (BRS)}
\label{subsec:balancoir}

\textit{Reservoir} sampling is an online update procedure that populates a fixed-size buffer with data coming from a stream. It guarantees each exemplar from that stream to be represented in the buffer with the same probability $\nicefrac{|\mathcal{B}|}{i}$ at any iteration $i$, which makes it equivalent to an offline random sampling at each time step. However, if the dataset $\mathcal{D}$ contains exemplars from $C$ distinct classes and we randomly sample $|\mathcal{B}|$ of them, the probability of leaving at least one class out is given by:
\begin{equation}
    P = \bigg(1 - \frac{1}{C}\bigg)^{|\mathcal{B}|}.
\end{equation}

This would result in leaving $1 / P$ classes out of $\mathcal{B}$, which becomes especially critical when dealing with small buffers: considering $|\mathcal{B}| \approx C$, that probability increases from $0.25$ for $C=2$ to $\approx 0.367$ for $C\to\infty$ (\textit{e.g.}\ $0.349$ for $C=10$).

To overcome such an issue, other approaches resort to the \textit{ring} buffer~\cite{chaudhry2019on} or the \textit{herding}~\cite{rebuffi2017icarl} strategy; however, these are not optimal in terms of buffer exploitation or computational overhead. The former reserves a slice as large as $\nicefrac{|\mathcal{B}|}{C}$ for each class: since classes are shown incrementally, this would leave the main part of the buffer empty. The latter changes the dimension of such slices with the number of seen classes, always reserving $\nicefrac{|\mathcal{B}|}{C_{seen}}$ slots for each class (where $C_{seen}$ indicates the number of seen classes). Despite its increased efficiency in terms of memory, \textit{herding} additionally performs a forward pass over the training set at the end of each task.

\begin{figure*}[t]
\centering
\includegraphics[width=.8\textwidth]{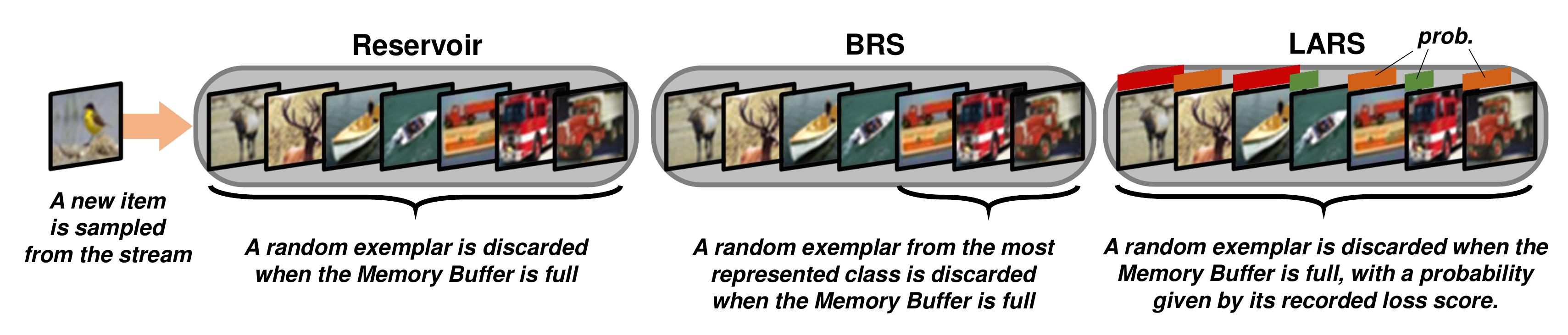}
\vspace{-1.0em}
\caption{Graphical comparison between reservoir, Balanced Reservoir and Loss-Aware Reservoir \textit{(best in color)}.}
\label{fig:coresets}
\end{figure*}

Instead, we propose Balanced Reservoir Sampling (see Alg.~\ref{alg:balancoir}), which encourages the balance within the buffer in terms of the number of exemplars per class. It consists in a small modification to \textit{reservoir}: instead of replacing a random exemplar when a newer one is inserted (line $8$ of Alg.~\ref{alg:balancoir}), we look for the element to be removed among those belonging to the most represented class (lines $9$-$11$). The difference is also graphically exemplified in Fig.~\ref{fig:coresets}.

\begin{figure}[t]
\centering
\includegraphics[width=0.48\textwidth]{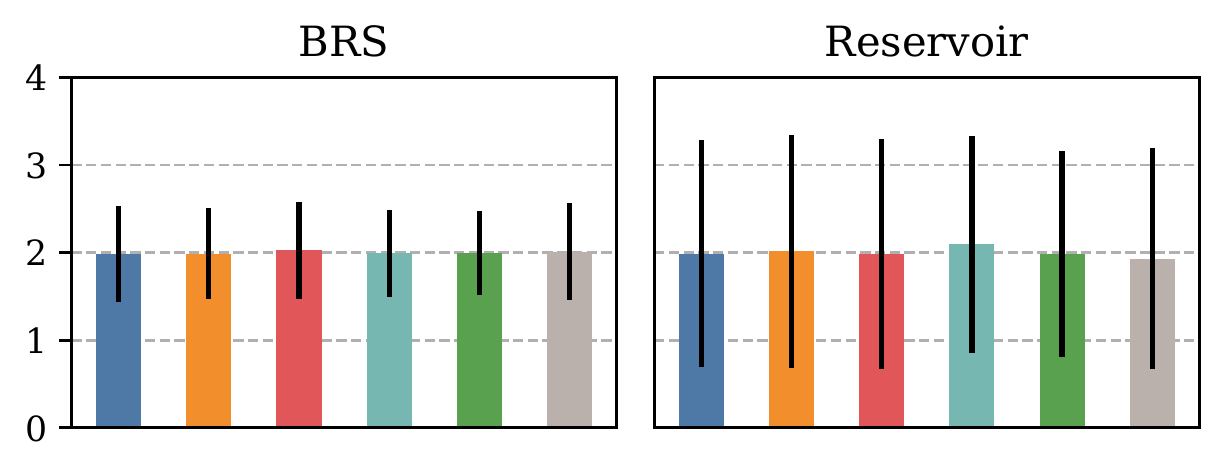}
\vspace{-0.5em}
\caption{Number of exemplars per class when applying different sampling strategies to the toy dataset described in Sec.~\ref{subsec:balancoir}, error bars indicate standard deviation. We consider a buffer size of $12$ items, with the objective of sampling exactly $2$ item per class \textit{(best in color)}.}
\vspace{-0.5em}
\label{fig:balance}
\end{figure}
To provide a better understanding, we compare Balanced Reservoir with \textit{reservoir} on a toy dataset of $1020$ items belonging to $6$ distinct classes ($170$ items per class). Fig.~\ref{fig:balance} shows the amount of samples per class retained at the end of each test. BRS is closer to the ideal solution of storing exactly $2$ items for each class than plain \textit{reservoir}. The two approaches achieve a Mean Squared Error of $0.28$ and $1.64$ respectively.

\subsection{Loss-Aware Reservoir Sampling (LARS)}
\label{subsec:lossoir}

\begin{algorithm}[t]
    \caption{Loss-Aware Balanced Reservoir Sampling}
    \label{alg:lossoir}
    \begin{algorithmic}[1]
      \STATE {\bfseries Input:} exemplar $(x,y,\ell)$, replay buffer $\mathcal{B}$, 
      \STATE \hspace{2.70em} number of seen examples $N$.
      \IF{$|\mathcal{B}| > N$}
            \STATE $\mathcal{B}[N]\gets (x,y, \ell)$
      \ELSE
            \STATE $j \gets \operatornamewithlimits{RandInt}([0, N])$
            \IF {$j < |\mathcal{B}|$}
                \STATE $ \boldsymbol{S_{balance}} \gets \{\operatornamewithlimits{ClassCounts}(y); \forall (x,y,\ell) \in \mathcal{B}\}$
                \STATE $ \boldsymbol{S_{loss}} \gets \{-\ell; \forall (x,y,\ell) \in \mathcal{B}\}$
                \STATE $ \alpha \gets \sum_k{|\boldsymbol{S_{balance}}[k]|} / \sum_k{|\boldsymbol{S_{loss}}[k]|}$
                \STATE $ \boldsymbol{S} \gets \boldsymbol{S_{loss}} \cdot \alpha + \boldsymbol{S_{balance}} $
                \STATE $ \boldsymbol{probs} \gets \boldsymbol{S} / \sum_k{\boldsymbol{S}[k]}$
                \STATE $k \gets \operatornamewithlimits{RandInt}([0, |\mathcal{B}|], \boldsymbol{probs})$
                \STATE $\mathcal{B}[k]\gets (x,y, \ell)$      
            \ENDIF
      \ENDIF
    \vspace{0.505em}
    \end{algorithmic}
\end{algorithm}

We introduce here an additional variation in the sampling strategy that limits the overfitting on buffer datapoints. Taking inspiration from~\cite{aljundi2019gradient}, we wish to make room for new examples by discarding those that are less important for preserving the performance. The authors of~\cite{aljundi2019gradient} devise both a rigorous Integer Quadratic Programming-based objective and a more efficient approximated greedy strategy for this purpose. However, since they resort to comparison between the gradients of individual examples, their proposal proves very slow w.r.t.\ to plain reservoir. Instead, we propose using the training loss value directly as a much simpler yet effective criterion for modeling the importance of examples. Indeed, the overall expected loss of the buffer can be computed without back-propagation and it should be maximized at all times, thus promoting the retention of exemplars that have not been fit.

To make \textit{reservoir} \textit{loss-aware}, we could identify and replace the elements displaying low loss values. These can be naively computed by feeding all the replay examples into the model before the replacement phase. However, this becomes computationally inefficient when the buffer is large, especially in earlier tasks when \textit{reservoir} replaces items more frequently. To overcome this issue, we propose an online update of the loss values: for every example that we store in the buffer, we also save the original loss score. As this is a scalar value, the memory overhead that results from storing it is negligible w.r.t.\ the cost of storing the example to be replayed. To keep the scores up-to-date, whenever the corresponding items are drawn for replay we replace the stored loss values with the current losses that are yielded by the model. 

Since they are complementary and address separate issues, we combine Loss-Aware Reservoir and BRS into a single algorithm (Alg.~\ref{alg:lossoir}). In doing so we: i) compute a $S_{balance}$ score vector proportional to the number of items of each class (line $8$); ii) estimate an importance score $S_{loss}$, given by the opposite of the loss value for each example (line $9$); iii) normalize these two terms to ensure an equal contribution and sum them to form a single score vector $S$ (lines $10$-$11$). Finally, we assign each a replacement probability to each item that is proportional to the combined score (lines $12$-$14$).

\input{tab2}

\section{Experimental Results}
\label{sec:exp}
\subsection{Experimental Protocol}

We test our proposal in the following Class-IL settings, characterized by an increasing difficulty:

\begin{itemize}
    \item Split Fashion-MNIST~\cite{chen2019overcoming} consists of five tasks featuring two classes each, with $6000$ exemplars per class. It is based on the Fashion-MNIST dataset~\cite{xiao2017fashion} which was designed as a drop-in replacement for MNIST~\cite{lecun1998gradient}. We rely on this benchmark instead of Split MNIST as the latter is simple and not representative of modern CV tasks~\cite{xiao2017fashion,farquhar2018towards};
    \item Like the previous one, Split CIFAR-10~\cite{zenke2017continual} is organized in five tasks with two classes each and $5000$ examples per class coming from the CIFAR-10 dataset~\cite{krizhevsky2009learning};
    \item Split CIFAR-100~\cite{zenke2017continual} consists of ten tasks, with ten distinct classes each and $500$ exemplars per class, deriving from the CIFAR-100 dataset~\cite{krizhevsky2009learning};
    \item Split CORe-50~\cite{maltoni2019continuous} is comprised of $50$ classes, with around $2400$ examples per class. In line with the SIT-NC protocol described in~\cite{maltoni2019continuous}, these classes are organized in nine tasks, the first of which includes ten classes whereas the following ones five each.
\end{itemize}

On the Split Fashion-MNIST setting, we apply all evaluated CL methods on a fully-connected network with two hidden layers of $256$ ReLU units each, in line with~\cite{lopez2017gradient, riemer2018learning}. On the other settings, we employ ResNet18~\cite{he2016deep} as a backbone in accordance to~\cite{rebuffi2017icarl}. All models are trained from scratch, iterating on each task for one epoch in Split Fashion-MNIST, $50$ epochs in Split CIFAR-10 and CIFAR-100 and $15$ epochs in Split CORe-50. We make use of Stochastic Gradient Descent (SGD) as optimizer.

To tune hyperparameters, we perform a grid-search on a validation set prior to training. This set is given by $6000$, $5000$ and $12000$ random items for the Fashion-MNIST, CIFAR-10/100 and CORe-50 datasets respectively. We keep the batch size fixed in order to guarantee an equal number of updates across all methods. All results in this section are expressed in terms of average accuracy over all tasks at the end of training. We conduct each test $10$ times and report the average result. We make the code for these experiments publicly available\footnote{\href{https://github.com/hastings24/rethinking_er}{https://github.com/hastings24/rethinking\_er}}.

\subsection{Comparison with State of the Art}
\label{subsec:stoa}

In this section, we draw a comparison between ER equipped with our tricks (ER+T) and state-of-the-art rehearsal methods, namely iCaRL~\cite{rebuffi2017icarl}, GEM~\cite{lopez2017gradient}, A-GEM~\cite{chaudhry2018efficient} and HAL~\cite{chaudhry2020using}. We test ER in combination with the \textit{reservoir} sampling strategy, while GEM, A-GEM, HAL use a \textit{ring} buffer and iCaRL leverages \textit{herding}. To provide a better understanding, Tab.~\ref{tab:results_competitors} reports the results for different buffer sizes ($200$, $500$, $1000$). We also train the backbones on the data-stream with no regularization (SGD) and on all shuffled datapoints (Joint Training), providing a lower and upper bound respectively.

The experiments on Split Fashion-MNIST show that ER+T consistently surpasses all competitors. Due to the mentioned weakness of \textit{reservoir}, the tricks prove to be especially beneficial when the memory buffer is smaller. However, plain Experience Replay is an already strong baseline, as revealed by GEM and A-GEM performing worse than it. HAL starts out on par with ER at reduced buffer size, but achieves weaker results when the latter increases. Interestingly, iCaRL only performs better than ER at memory size $200$. This is due to the \textit{herding} strategy, through which the method fills the buffer with the best possible exemplars for its classification procedure. By so doing, iCaRL gains an advantage over methods applying \textit{reservoir} and \textit{ring} that is less prominent for larger buffer sizes.

Experiments on the harder Split CIFAR-10 and Split CIFAR-100 protocols show GEM and iCaRL prevailing over naive ER. In particular, iCaRL sets a very high bar on hard datasets thanks to its simple and effective nearest-mean-of-exemplars classification rule. Nevertheless, the application of our proposed tricks gives ER+T an edge over the competition. Notably, HAL encounters some failures on Split CIFAR-100.

Even on CORe-50, the tricks yield a boost in performance compared to naive replay, outperforming the other approaches. In contrast with what we have observed for Split CIFAR-10 and -100, iCaRL does not achieve reliable performance. We ascribe this finding to the nature of CORe-50 in Class-IL: this dataset shows very similar entities (\textit{e.g.}: slightly different plug adapters) as separate classes. While such subtle differences are successfully learned by ER through backpropagation, iCaRL strictly depends on its nearest-mean-of-exemplars classifier, which can be in trouble in distinguishing fine-grained details.

\subsection{Influence of Each Trick}
\input{tab1}
\input{tab3}
To quantify the effect of each trick presented in Sec.~\ref{sec:tricks}, we apply them increasingly to ER (Tab.~\ref{tab:incr_results}, Fig.~\ref{fig:perf_tricks}). Since we do not apply data augmentation to the input batches of Fashion-MNIST, we do not employ it on buffer points either. 
IBA proves to be very effective on both the CIFAR-10 and CIFAR-100 datasets, providing a very meaningful boost in accuracy when compared to the initial performance. This turns out to be especially remarkable in the former setting, where it almost doubles the initial accuracy. Both Bias Control and Exponential learning rate Decay present a solid positive effect, especially when looking at the two most difficult settings. Finally, the combination of the two proposed sampling strategies (BRS and LARS) is particularly beneficial in the context of the Fashion-MNIST and CIFAR-10 experiments.

Overall, we observe a remarkable performance boost in CIFAR-10 and CIFAR-100 ($+146\%$ and $+120\%$ respectively). This shows that the proposed tricks are very effective in challenging datasets where ER can still improve.

\subsection{Applicability of IBA to other rehearsal methods}
\label{subsec:daug}
\begin{table}[t]
\centering
\caption{Accuracy values on Split Fashion-MNIST when applying the proposed tricks by the use of a $500$-examples buffer.}
\setlength\tabcolsep{8pt}
\begin{tabular}{lcccc}  
\toprule
\textbf{Methods} & \textbf{No trick} & \textbf{BiC} & \textbf{CBiC} & \textbf{CBiC+ElrD}\\
\midrule
SI       & $19.91$ & $24.67$ & $33.15$ & $35.51$ \\
oEWC     & $20.04$ & $25.71$ & $40.36$ & $43.85$ \\
\bottomrule
\end{tabular}
\label{tab:cbic_results}
\end{table}
\begin{figure}[t]
\includegraphics[width=0.475\textwidth]{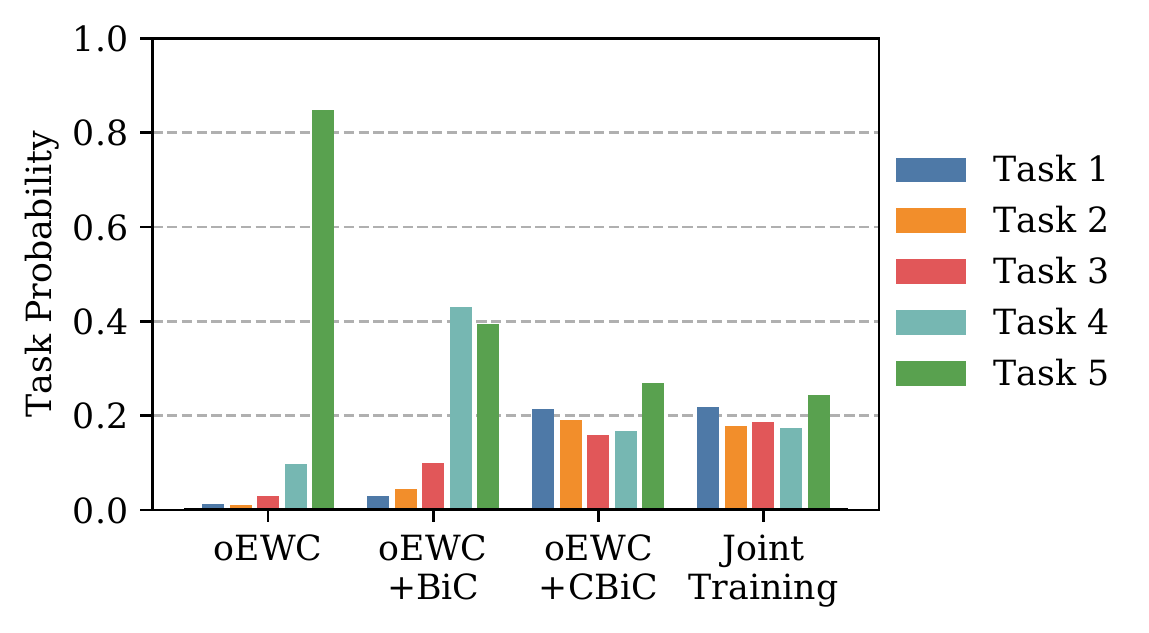}
\captionof{figure}{Probability for each task to be predicted at the end of the training on Split Fashion-MNIST, for oEWC with different bias correction layers and Joint Training. We apply BiC and CBiC to oEWC by training it on a \textit{reservoir} memory buffer included for the purpose \textit{(best in color)}.}
\label{fig:tmr}
\end{figure}
Data augmentation is a very simple and effective technique allowing to virtually draw from a larger amount of data. As it often comes for free (no additional costs in terms of annotations or storage are required), the more a method can benefit from it, the better. 
This especially holds for rehearsal-based methods approaching CL scenarios, where only few examples are available to guard the old knowledge. Surprisingly, we find out that not all methods benefit from data-augmentation on buffer datapoints (IBA). In this respect, Tab.~\ref{tab:results_daug} shows that there is not a clear trend: while HAL always improves its performance in a consistent way, GEM suffers from a severe degradation on Split CIFAR-10. We conjecture that this is due to the way GEM makes use of the buffer: indeed, satisfying the inequality constraints given by augmented examples could not link well to retaining the performance on the original tasks.

\subsection{Applicability to non-Rehearsal Methods}
\label{subsec:perf_non_repl}
We show here that Bias Control (BiC) and Exponential learning rate Decay (ElrD) can be extended to two prior-based approaches, namely EWC online (oEWC)~\cite{schwarz2018progress} and Synaptic Intelligence (SI)~\cite{zenke2017continual}. The latter anchor most important parameters at the optimal point when the task changes, yielding remarkable performance when task-label is provided at inference time. However, when dealing with the Class-IL scenario, they completely forget all tasks but the latest. As observed in~\cite{wu2019large}, this is mainly due to an implicit bias induced by the optimization performed on the current task.

It is worth noting that BiC compensates the responses related to the last task solely. Consequently, it proves to be effective only under the hypothesis that previous tasks are equally biased. This holds for Experience Replay and other rehearsal methods, which rehearse an equal amount of exemplars from the previous tasks at each iteration. Conversely, oEWC and SI do not access past examples at all, which discourages the classifier from predicting older tasks. To shed light on these issues, Fig.~\ref{fig:tmr} shows the probability for each task of being predicted, obtained as follows: i) for each test example, we compute the output distribution over the classes; ii) we average the probabilities over all exemplars task-wise; iii) we normalize the results to describe them as probability distributions. As can be seen, BiC effectively reduces the bias displayed by oEWC towards the last task. However, since the model also shows a tendency to predict classes of earlier tasks, BiC cannot manage to balance out all prior probabilities.
\input{fig3}
On these grounds, we adopt here a tailored variant for oEWC and SI, namely Complete Bias Correction (CBiC)\footnote{It must be noted that, since Experience Replay is evenly biased w.r.t.\ previous tasks, BiC and CBiC are equally effective on it. For this reason, we advise sticking to the former for ER as it is simpler and thus easier to train.}. Whilst BiC corrects the logits related to the latest task solely, CBiC adjusts the responses related to each task independently. Technically, CBiC applies an additive offset $\beta_t$ to the logits related to each task $t$. Fig.~\ref{fig:tmr} shows that this approach favorably results in a flatter distribution for oEWC, which is closer to the one of the Joint Training. This is confirmed by the accuracy results in Tab.~\ref{tab:cbic_results}, which also illustrates how the additional application of ElrD further increases its performance. 

\subsection{Execution Times}

To further account for the differences among the listed methods, we illustrate their run times in Fig.~\ref{fig:times}. This experiment refers to the Split Fashion-MNIST dataset, employing a memory buffer of size $200$. We ran tests on identical conditions for all methods, using a NVidia GeForce RTX 2080 GPU.

Thanks to its simplicity, plain ER is remarkably faster than all other methods. On the other side, GEM is by far the slowest method. As it relies on a demanding quadratic programming constraint for each task, its wallclock time increases remarkably as the training progresses. By comparison, A-GEM is clearly much faster as it only applies one similar constraint at all times. iCaRL, HAL and ER+T have similar medium-long execution times linked to computations at task boundaries. In this phase, the \textit{herding} strategy of iCaRL examines all examples of the previous task, HAL computes one anchor point per class and ER+T trains the Bias Control module. Another factor slowing the latter method down is the increased amount of computation required by BRS and LARS. As \textit{reservoir} samples more frequently at the beginning of training, this effect fades in subsequent tasks.

\section{Conclusions}
In this work, we contribute to Continual Learning with a collection of tricks enhancing Experience Replay. We address the Class Incremental Learning setting as it is the most challenging among the standard CL scenarios. To show the effectiveness of our proposals, we conduct a comparison among state-of-the-art rehearsal methods on increasingly harder datasets. Despite a limited growth in computational requirements, ER equipped with our tricks outperforms more sophisticated approaches. Finally, we show that some of our proposals can be beneficially applied even to prior-based methods. 
\bibliographystyle{IEEEtran}
\balance
\bibliography{references}

\end{document}

%% file: header.tex
\usepackage{lipsum}
\usepackage{xcolor}
\usepackage{epigraph}
\usepackage{graphicx}
\usepackage{amsmath}
\usepackage{booktabs}
\usepackage{multicol}
\usepackage{amssymb}
\usepackage{multirow}
\usepackage{dsfont}
\usepackage{balance}
\usepackage[first=0, last=9]{lcg}
\usepackage{algorithm}
\usepackage{algorithmic}
\usepackage{enumitem}
\usepackage[normalem]{ulem}
\usepackage{nicefrac}
\usepackage{capt-of}
\usepackage{url}
\usepackage{comment}
\usepackage{tabularx}
\usepackage{hyperref}

\newcommand{\argmax}{\operatornamewithlimits{argmax}}

\newcommand{\xmark}{\textbf{\textendash}}

\usepackage{twoopt}
\usepackage{tikz}

\usetikzlibrary{fit}

\newcommand\tikzmark[1]{%
  \tikz[remember picture,overlay]\node[inner xsep=0pt] (#1) {};}

\newcommandtwoopt\TextboxReservoir[5][2.5cm][2cm]{%
\begin{tikzpicture}[remember picture,overlay]
  \coordinate (aux) at ([xshift=#1]#4);
  \node[inner ysep=5pt,yshift=0.6ex,draw=black,thick,
    fit=(#3) (aux),baseline] 
    (box) {};
\end{tikzpicture}%
}
\newcommandtwoopt\TextboxBalancedReservoir[5][2.5cm][2cm]{%
\begin{tikzpicture}[remember picture,overlay]
  \coordinate (aux) at ([xshift=#1]#4);
  \node[inner ysep=17pt,yshift=0.5ex,draw=black,thick,
    fit=(#3) (aux),baseline] 
    (box) {};
\end{tikzpicture}%
}

%% file: abs.tex
In Continual Learning, a Neural Network is trained on a stream of data whose distribution shifts over time. Under these assumptions, it is especially challenging to improve on classes appearing later in the stream while remaining accurate on previous ones. This is due to the infamous problem of catastrophic forgetting, which causes a quick performance degradation when the classifier focuses on learning new categories. Recent literature proposed various approaches to tackle this issue, often resorting to very sophisticated techniques. In this work, we show that naive rehearsal can be patched to achieve similar performance. We point out some shortcomings that restrain Experience Replay (ER) and propose five tricks to mitigate them. Experiments show that ER, thus enhanced, displays an accuracy gain of $\boldsymbol{51.2}$ and $\boldsymbol{26.9}$ percentage points on the CIFAR-10 and CIFAR-100 datasets respectively (memory buffer size $\boldsymbol{1000}$). As a result, it surpasses current state-of-the-art rehearsal-based methods.

%% file: tab2.tex
\begin{table*}[t]
\small
\centering
\caption{Comparison among state-of-the-art methods in terms of average final accuracy on several datasets.}
\setlength\tabcolsep{5pt}
\begin{tabular}{lcccccccccccc}  
\toprule
\textbf{Methods} & \multicolumn{3}{c}{\textbf{Split Fashion-MNIST}} & \multicolumn{3}{c}{\textbf{Split CIFAR-10}} & \multicolumn{3}{c}{\textbf{Split CIFAR-100}} & \multicolumn{3}{c}{\textbf{Split CORe-50}}\\
\midrule
SGD             & \multicolumn{3}{c}{$20.11$}    & \multicolumn{3}{c}{$19.62$} &   \multicolumn{3}{c}{$8.54$}  & \multicolumn{3}{c}{$8.89$}\\
Joint Training  & \multicolumn{3}{c}{$84.47$}    & \multicolumn{3}{c}{$92.13$} &   \multicolumn{3}{c}{$70.66$} & \multicolumn{3}{c}{$49.51$} \\
\midrule
\textbf{Memory Buffer Size} & $\mathcal{B}_{200}$ & $\mathcal{B}_{500}$ & $\mathcal{B}_{1000}$ & 
   $\mathcal{B}_{200}$ & $\mathcal{B}_{500}$ & $\mathcal{B}_{1000}$ & 
   $\mathcal{B}_{200}$ & $\mathcal{B}_{500}$ & $\mathcal{B}_{1000}$ & 
   $\mathcal{B}_{200}$ & $\mathcal{B}_{500}$ & $\mathcal{B}_{1000}$\\
\midrule
A-GEM~\cite{chaudhry2018efficient}  & $49.73$    & $49.47$  & $50.98$ & $19.90$  & $20.35$  & $19.81$ & $9.17$  & $9.23$   & $9.12$  & $9.33$  & $9.42$  & $8.96$ \\
GEM~\cite{lopez2017gradient}        & $69.46$    & $75.91$  & $79.62$ & $28.14$  & $34.69$  & $36.68$ & $9.18$  & $14.12$  & $17.88$ & \xmark  & \xmark  & \xmark \\
HAL~\cite{chaudhry2020using}        & $72.59$    & $77.59$  & $80.79$ & $25.92$  & $27.99$  & $29.10$ & $7.63$  & $9.66$  & $10.43$ & $11.53$  & $12.40$  & $8.53$ \\
iCaRL~\cite{rebuffi2017icarl}       & $75.46$    & $77.54$  & $78.13$ & $41.26$  & $41.34$  & $42.03$ & $20.73$ & $24.74$  & $25.52$ & $8.01$  & $7.23$  & $8.05$ \\
ER~\cite{ratcliff1990connectionist} & $72.54$    & $79.02$  & $81.39$ & $24.06$  & $27.06$  & $31.38$ & $9.66$  & $11.50$  & $12.36$ & $19.48$ & $28.54$ & $32.66$ \\
ER+T (ours)  & $\boldsymbol{76.07}$  & $\boldsymbol{80.11}$  & $\boldsymbol{82.46}$ 
             & $\boldsymbol{59.18}$  & $\boldsymbol{62.60}$  & $\boldsymbol{70.99}$ 
             & $\boldsymbol{21.26}$  & $\boldsymbol{24.90}$  & $\boldsymbol{36.05}$  
             & $\boldsymbol{25.63}$ & $\boldsymbol{33.33}$ & $\boldsymbol{37.44}$ 
             \\
\bottomrule
\end{tabular}
\label{tab:results_competitors}
\end{table*}

%% file: tab1.tex
\begin{table}[t]
\small
\centering
\caption{Accuracy values on several datasets as more tricks from Sec~\ref{sec:tricks} are added to the baseline, with replay buffer size $200$.}
\setlength\tabcolsep{2pt}
\begin{tabular}{lccc}  
\toprule
\multirow{2}{*}{\textbf{Methods}} & \textbf{Split} & \textbf{Split} & \textbf{Split}\\
 & \textbf{Fash-MNIST} & \textbf{CIFAR-10} & \textbf{CIFAR-100}\\
\midrule
ER                  & $72.54$ & $24.06$  & $ 9.66$ \\
~+ IBA              & \xmark  & $44.78$  & $13.90$ \\
~+ BiC              & $73.43$ & $49.27$  & $17.73$ \\
~+ ElrD             & $74.19$ & $51.02$  & $20.27$ \\
~+ BRS              & $74.66$ & $52.75$  & $20.64$ \\
~+ LARS             & $76.07$ & $59.18$  & $21.26$ \\
\bottomrule
\end{tabular}
\label{tab:incr_results}
\end{table}

%% file: tab3.tex
\begin{table}[t]
\small
\centering
\caption{Ablation study inquiring the impact of IBA (Sec.~\ref{subsec:mem_aug}).}
\setlength\tabcolsep{4pt}
\begin{tabular}{lcccccc}  
\toprule
\textbf{Methods} & \multicolumn{3}{c}{\textbf{Split CIFAR-10}} & \multicolumn{3}{c}{\textbf{Split CIFAR-100}}\\
\midrule
\textbf{Memory Buffer} & $\mathcal{B}_{200}$ & $\mathcal{B}_{500}$ & $\mathcal{B}_{1000}$ & 
   $\mathcal{B}_{200}$ & $\mathcal{B}_{500}$ & $\mathcal{B}_{1000}$\\
\midrule
A-GEM~\cite{chaudhry2018efficient} & $19.90$  & $20.35$ & $19.81$ & $9.17$ & $9.23$ & $9.12$ \\
~+ IBA & $20.23$  & $19.97$ & $21.15$ & $9.16$  & $9.34$ & $9.39$   \\
\midrule
GEM~\cite{lopez2017gradient} & $28.14$  & $34.69$  & $36.68$ & $9.18$  & $14.12$  & $17.88$ \\
~+ IBA        & $22.62$  & $23.01$ & $20.25$   & $13.69$  & $16.74$ & $15.21$  \\
\midrule
HAL~\cite{chaudhry2020using} & $25.92$  & $27.99$  & $29.10$ & $7.63$  & $9.66$  & $10.43$ \\
~+ IBA        & $32.33$  & $41.77$ & $49.28$ & $8.19$  & $11.39$ & $12.91$ \\
\bottomrule
\end{tabular}
\label{tab:results_daug}
\end{table}

%% file: fig3.tex
\begin{figure}[t]
\includegraphics[width=0.475\textwidth]{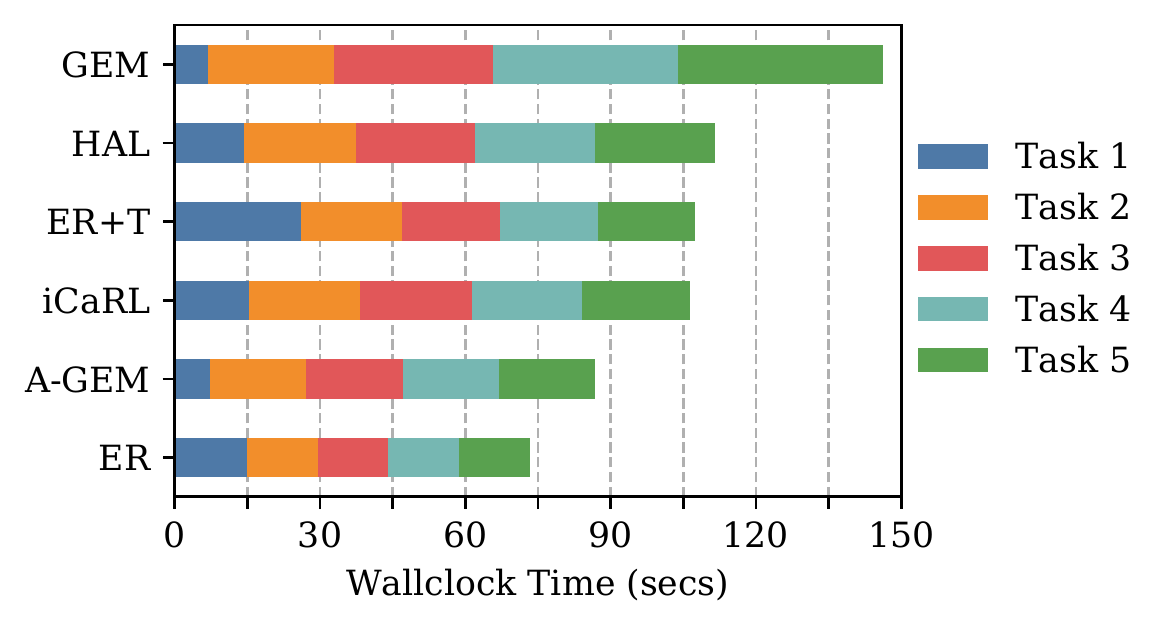}
\captionof{figure}{Run times for different methods on the Split Fashion-MNIST setting, with buffer size $200$ \textit{(best in color)}.}
\label{fig:times}
\end{figure}